%% file: example_paper.tex
\documentclass{article}

\usepackage{microtype}
\usepackage{graphicx}
\usepackage{subfigure}
\usepackage{booktabs} 

\usepackage{hyperref}

\usepackage[accepted]{icml2025}

\usepackage{amsmath}
\usepackage{amssymb}
\usepackage{mathtools}
\usepackage{amsthm}
\usepackage{times}
\usepackage{latexsym}
\usepackage[T1]{fontenc}
\usepackage[utf8]{inputenc}
\usepackage[utf8]{inputenc}
\usepackage{newunicodechar}

\newunicodechar{【}{[}
\newunicodechar{】}{]}
\usepackage{caption} 
\usepackage{tablefootnote}
\usepackage{microtype} 
\usepackage{inconsolata}
\usepackage{graphicx}
\usepackage{enumitem}
\usepackage{booktabs} 
\usepackage{soul}
\usepackage{xcolor,colortbl} 
\usepackage{changepage,threeparttable} 
\usepackage{amsmath}
\usepackage{pifont}
\usepackage{bm}
\usepackage{amssymb}
\usepackage{amsfonts}
\usepackage{adjustbox}
\usepackage{multirow}
\usepackage{algorithm}
\usepackage{algorithmic}
\usepackage{ragged2e}
\usepackage{makecell}

\usepackage[capitalize,noabbrev]{cleveref}

\usepackage[textsize=tiny]{todonotes}

\icmltitlerunning{AdaRec: Adaptive Recommendation with LLMs via Narrative Profiling and Dual-Channel Reasoning}

\begin{document}

\twocolumn[
\icmltitle{AdaRec: Adaptive Recommendation with LLMs via \\ Narrative Profiling and Dual-Channel Reasoning} 

\begin{icmlauthorlist}
\icmlauthor{Meiyun Wang}{yyy}
\icmlauthor{Charin Polpanumas}{comp}
\end{icmlauthorlist}

\icmlaffiliation{yyy}{The University of Tokyo}
\icmlaffiliation{comp}{Amazon.com, Inc.}

\icmlcorrespondingauthor{Meiyun Wang}{omiun20@g.ecc.u-tokyo.ac.jp}


\vskip 0.3in
]

\printAffiliationsAndNotice{}  

\begin{abstract} We propose \textbf{AdaRec}, a few-shot in-context learning framework that leverages Large Language Models (LLMs) for an adaptive personalized recommendation. AdaRec introduces \textbf{narrative profiling}, transforming user-item interactions into natural language representations to enable unified task handling and enhance human readability. Centered on a \textbf{bivariate reasoning paradigm}, AdaRec employs a \textbf{dual-channel architecture} that integrates \textbf{horizontal behavioral alignment}—discovering peer-driven patterns—with \textbf{vertical causal attribution}—highlighting decisive factors behind user preferences. Unlike existing LLM-based approaches, AdaRec \textbf{eliminates manual feature engineering} through semantic representations and supports \textbf{rapid cross-task adaptation} with minimal supervision. Experiments on real e-commerce datasets demonstrate that AdaRec outperforms both machine learning models and LLM-based baselines by up to \textbf{8\%} in few-shot settings. In zero-shot scenarios, it achieves up to a \textbf{19\%} improvement over expert-crafted profiling, showing effectiveness for long-tail personalization with minimal interaction data. Moreover, lightweight fine-tuning on synthetic data generated by AdaRec matches the performance of fully fine-tuned models, highlighting its efficiency and generalization across diverse tasks.\footnote{Codes are available: \url{https://anonymous.4open.science/r/AdaRec-CE5C}}. \end{abstract}

\input{docs/1_introduction}
\input{docs/2_relatedwork}
\input{docs/3_method}

\input{docs/4_experiment}

\input{docs/5_conclusion}

\nocite{langley00}

\bibliography{custom, anthology}
\bibliographystyle{icml2025}

\newpage
\appendix
\onecolumn
\section{Prompt Template for AdaRec}

\begin{table}[h]
  \centering
  \small
  \begin{tabular}{p{\linewidth}}
    \toprule
    \textbf{System Prompt} \\
    You are a customer profile generator. Below is the data distribution for each feature:\\
    \texttt{[Textual Distribution]}\tablefootnote{See the code for full details of [Textual Distribution].}\\  
    'Number of category purchased in the last 360 days.' has a mean value of 11.2 with a standard deviation of 6.6. The minimum observed value is 0.0, while the maximum is 34.0. Approximately 25\% of values are below 6.0, the median (50th percentile) is 12.0, and 75\% fall below 16.0 \dots{}. \\[1ex]
    Using this information, generate a clear and cohesive profile for the customer. For non-numerical features, emphasize specific values. For numerical features, describe relative trends without exact numbers. Present as a single fluid paragraph without extra formatting. Customer Profile:  \\[1ex]
    \textbf{Customer Profile}\\  
    Number of category purchased in the last 360 days is 6. Number of category viewed in the last 30 days is 3. Number of benefits used in the last 360 days is 4 \dots{} \\ 
    \bottomrule
  \end{tabular}
  \caption{Example of narrative profiling prompt.}
  \label{tbl:dist}
\end{table}

\begin{table}[h]
  \centering
  \small
  \begin{tabular}{p{\linewidth}}
    \toprule
    \textbf{System Prompt} \\
    As the Senior Marketing Manager at [company], your task is to recommend three brands for the promotional carousel. The promotion has two components: 1. Condition: Customers must purchase X units. 2. Reward: Customers receive 10\% in Points. \\[1ex]
    \texttt{[brand description]} \\
    Available brands: \\
    cocacola: Grocery beverages \dots{}\\[1ex]
    Based on a customer profile, please recommend three brand names for the customer. \\[1ex]
    \textbf{Factor Analysis} \\
    Reference: Factors and their importance ranking that affect brand recommendations based on historical data. \\
    \texttt{[Factor Analysis Guidelines]}\tablefootnote{Full versions of [Factor Analysis Guidelines] and [Pattern Analysis Guidelines] are available in the codes.} \\[1ex]
    \textbf{Pattern Analysis} \\
    Reference: Below are preferences from similar customer profiles. \\
    \texttt{[Pattern Analysis Guidelines]} \\[1ex]
    Reference Cases: \\
    Customer Profile: Number of days visited in the last 360 days is 315, Number of category viewed in the last 30 days is 5, Total mobile app visits in the last 360 days is 522 \dots{} \\[1ex]
    Based on the information above, please recommend three brand names in the following JSON format: \texttt{\{'brand': 'brand1, brand2, brand3', 'confidence': confidence, 'reason': reason\}} \\[1ex]
    \textbf{Narrative Profile} \\
    This customer is an active Prime member ... focusing primarily on wireless products and sports items... \\
    \bottomrule
  \end{tabular}
  \caption{Example of the structured reasoning prompt.}
  \label{tbl:brand_recommendation}
\end{table}

\section{Comparison of Expert Profiling and Narrative Profiling}
\begin{figure}[t]
\centering
\includegraphics[width=0.95\linewidth]{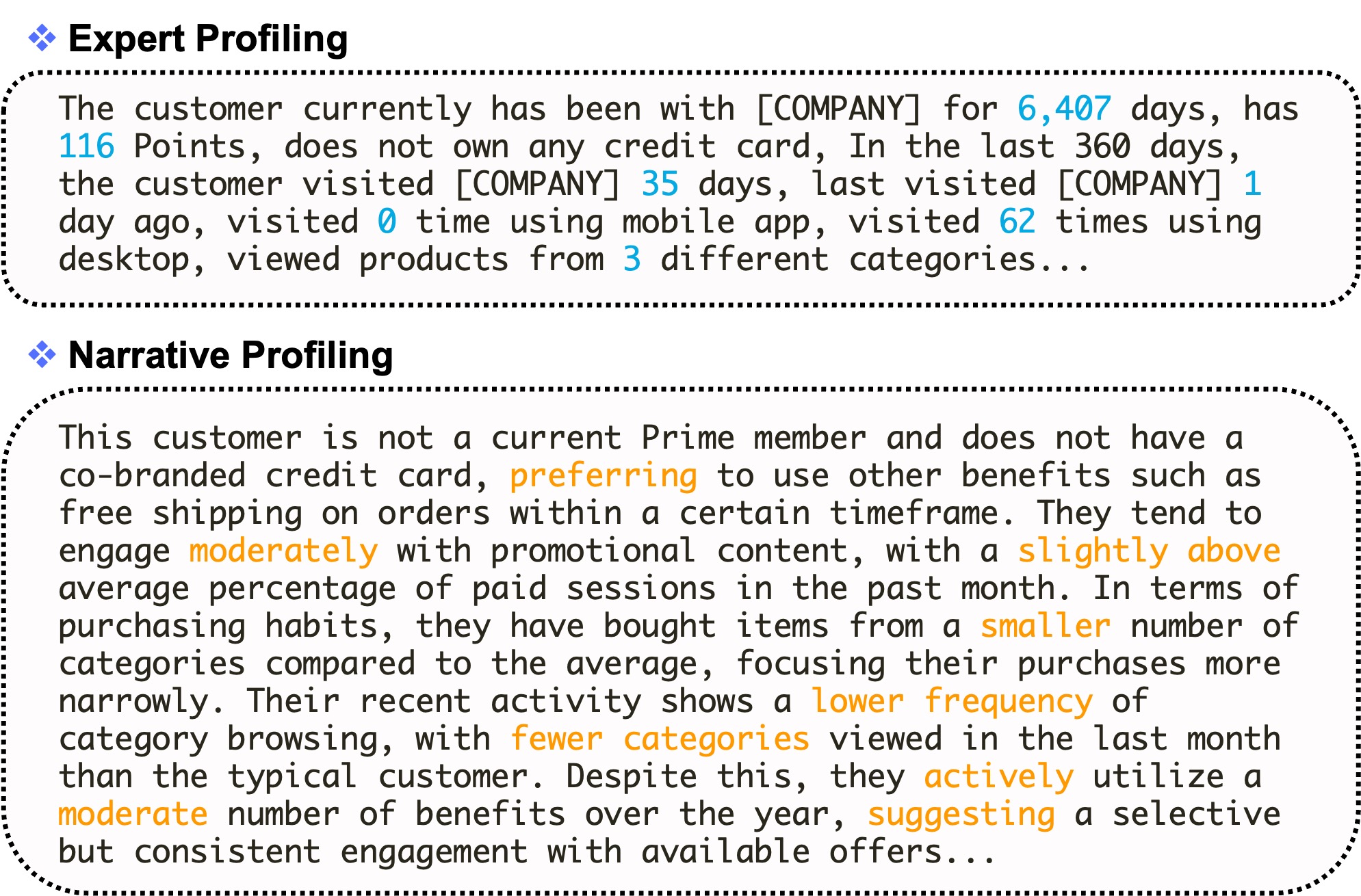}
\caption{Expert Profiling vs. Narrative Profiling.}
\label{fig:profile_comparison}
\end{figure}

Figure \ref{fig:profile_comparison} compares expert profiling and narrative profiling.

\section{Case Study}
\begin{figure}[h]
\centering
\includegraphics[width=\linewidth]{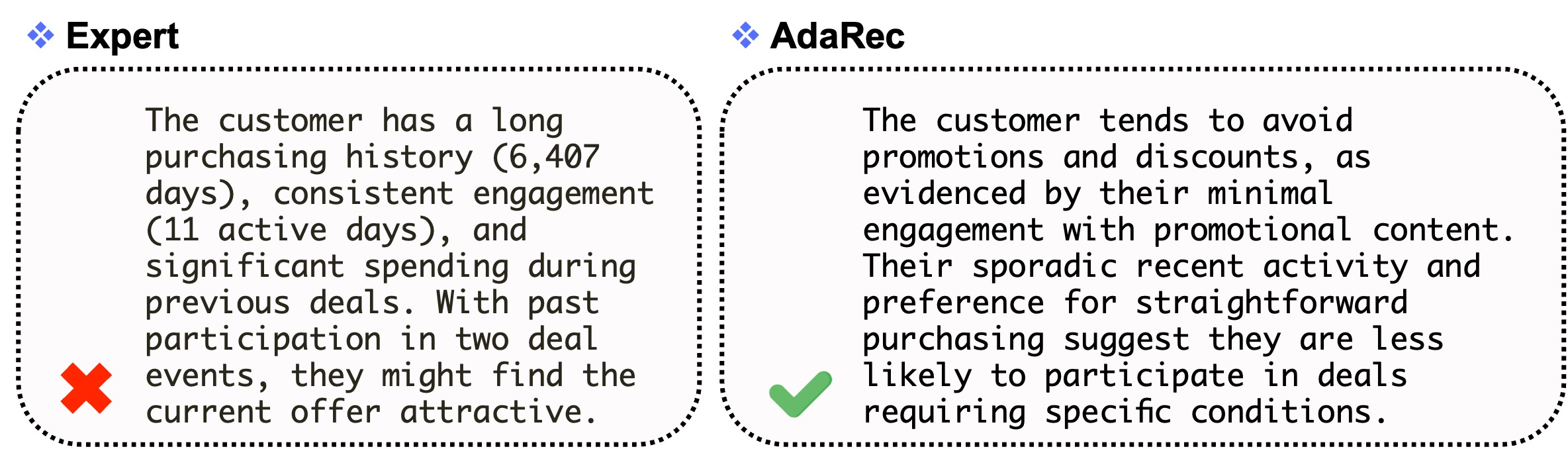}
\caption{AdaRec vs. expert analysis: capturing behavioral patterns for explainable recommendations.}
\label{fig:case}
\end{figure}

Figure~\ref{fig:case} compares expert reasoning with AdaRec. While experts rely on descriptive statistics, AdaRec captures nuanced behavioral patterns and underlying customer preferences, offering both accurate predictions and explainable insights.

\end{document}

%% file: docs/1_introduction.tex
\section{Introduction}
Recommender systems are critical to e-commerce, social media, and digital services, where delivering personalized content demands continuous adaptation to dynamic user preferences and evolving contexts \cite{hansen2020contextual}. Traditional approaches, including collaborative filtering \cite{he2017neural} and feature-based machine learning models \cite{weng2004feature, wang2015collaborative}, have achieved early success but struggle with manual feature engineering, limited generalization, and inflexibility in responding to changing user behaviors \cite{hu-etal-2020-graph}.

\begin{figure}[t]
\centering
\includegraphics[width=\linewidth]{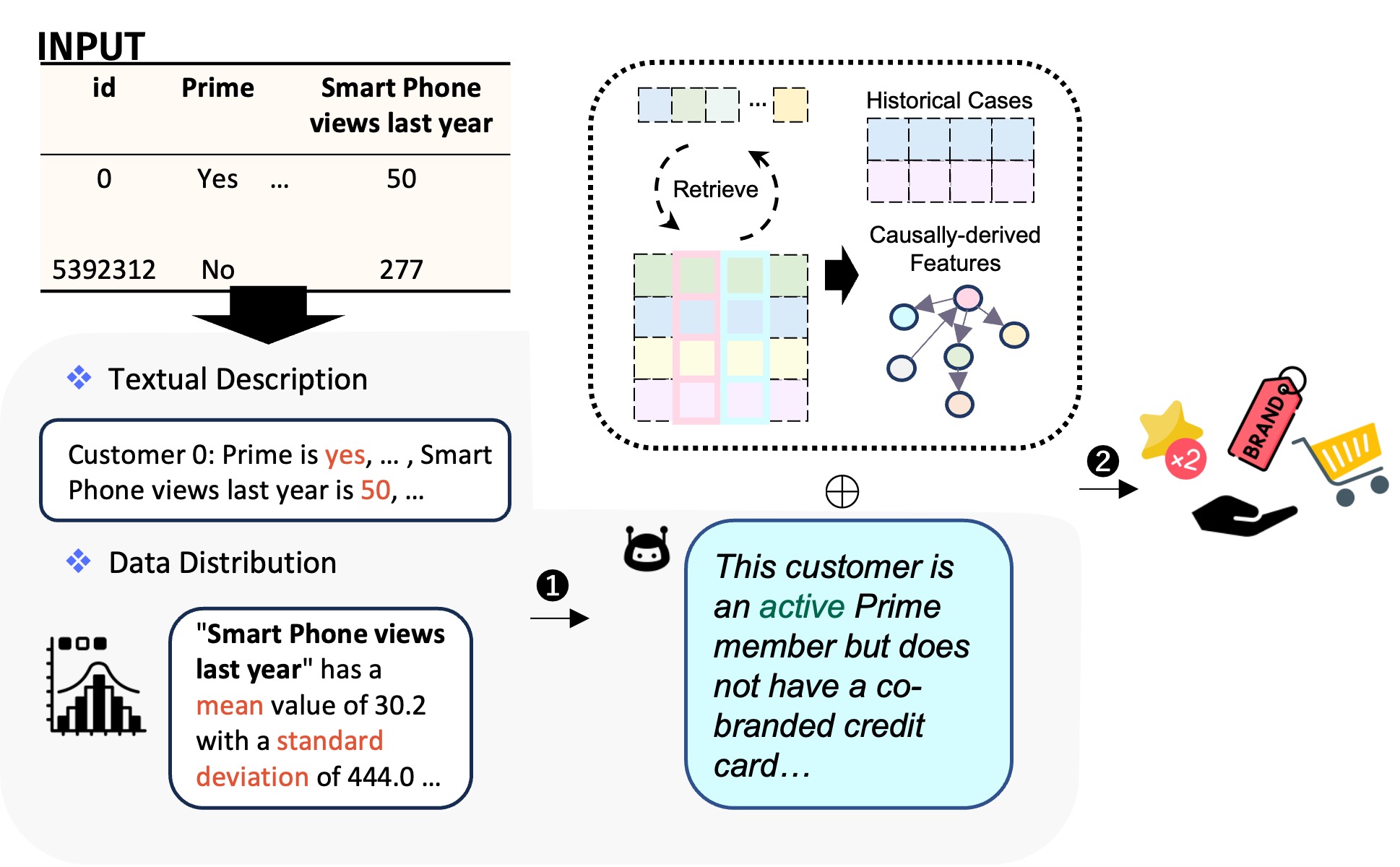}
\vspace{-0.08in}
\caption{AdaRec: Few-shot adaptive recommendation with LLMs via narrative profiling and dual-channel reasoning.}
\label{fig:intro}
\vspace{-2em}
\end{figure}

The rise of Large Language Models (LLMs) introduces new opportunities for recommender systems by leveraging semantic understanding and in-context learning capabilities \cite{wu2024survey}. Current LLM-driven recommendation research follows four main directions:  narrative-based methods generate static textual profiles but lack dynamic adaptability to evolving user preferences \cite{mysore2023large}; similarity-based approaches retrieve historical cases without causal reasoning, limiting robustness to data shifts \cite{xie2024neighborhood}; agent-based frameworks improve personalization via interaction but incur high operational costs \cite{wang2023recmind}; and methods such as fine-tuning-based techniques, multi-modal personalization systems, and feedback-driven exploration strategies \cite{lin2024data, zhang2024notellm} offer strong task-specific performance by tightly integrating LLMs into recommendation workflows. However, these methods rely on substantial computational resources and complex training pipelines. This highlights the need for a \textbf{lightweight}, \textbf{adaptable}, and \textbf{explainable} solution that minimizes task-specific 
 engineering overhead while maintaining high recommendation quality.
 
To address these challenges, we propose \textbf{AdaRec}, an LLM-based few-shot recommendation framework for adaptive personalization in tabular data. AdaRec introduces \textbf{narrative profiling}, converting user-item interactions into natural language to enhance flexibility and human-readable reasoning. It further employs a \textbf{dual-channel reasoning} architecture, integrating \textbf{horizontal behavioral alignment} with \textbf{vertical causal attribution} to combine peer pattern discovery and dynamic feature focus for robust, explainable recommendations.

The key contributions of this work are:
(1) \textbf{Efficient zero-shot and few-shot adaptation}, eliminating costly fine-tuning and manual feature engineering;  
(2) \textbf{Enhanced robustness and explainability} through integrated causal reasoning and narrative profiling;  
(3) \textbf{Seamless cross-task generalization}, enabling rapid deployment across diverse recommendation scenarios.

Experiments on real-world e-commerce datasets show that AdaRec consistently outperforms traditional machine learning models and recent LLM-based baselines, achieving up to \textbf{8\%} improvement on benchmark datasets and up to \textbf{19\%} gains in zero-shot settings over expert-crafted profiling. These results demonstrate that AdaRec is a practical and adaptable solution for recommendation challenges, balancing high performance with reduced manual effort and improved task flexibility.

%% file: docs/2_relatedwork.tex
\section{Related Work}

Before the rise of Large Language Models (LLMs), recommendation systems for tabular data focused on explicit user behavior modeling and feature interaction design. Representative methods include DIN \cite{zhou2018deep} for adaptive interest extraction, DIEN \cite{zhou2019deep} for temporal preference modeling, MIMN \cite{pi2019practice} and SIM \cite{pi2020search} for long-term behavior handling, and STAR \cite{sheng2021one} for multi-domain adaptation. While effective, these approaches rely heavily on manual feature engineering and task-specific architectures, limiting flexibility and explainability.

LLMs introduce new opportunities by addressing data sparsity, enhancing interaction via natural language, and offering strong generalization without complex feature processing. Fine-tuning methods like InstructRec \cite{zhang2023recommendation} and TALLRec \cite{bao2023tallrec} improve task-specific performance but demand high computational costs and retraining. Lightweight solutions such as RecMind \cite{wang2023recmind}, narrative-driven frameworks \cite{mysore2023large}, and similarity-based recommenders \cite{xie2024neighborhood} leverage in-context learning (ICL) and external knowledge but lack explainability. AdaRec addresses these gaps by combining LLM-driven flexibility and explainability with dual-channel reasoning, enabling lightweight, adaptive recommendations for tabular data without extensive fine-tuning or manual intervention.

%% file: docs/3_method.tex
\section{AdaRec}

AdaRec is a lightweight in-context learning framework that transforms tabular user data into natural language profiles via narrative profiling, enhancing recommendations through structured reasoning. By combining similarity retrieval and causal inference, AdaRec delivers personalized and explainable recommendations.

\textbf{Narrative Profiling.}  
For each user feature vector $\mathbf{x}_\theta \in \mathbb{R}^d$, AdaRec leverages LLMs to generate context-aware natural language descriptions. Instead of predefined rules, AdaRec provides global statistical distributions and user-specific data via carefully designed prompts. The LLM interprets each feature $f_i$ with value $x_{\theta,i}$ based on its distribution $\mathcal{D}_i$, generating context-aware qualitative descriptions: $\Psi(f_i, x_{\theta,i}, \mathcal{D}_i) \rightarrow \mathcal{T}$, where $\mathcal{T}$ denotes the natural language description space, and $\Psi$ represents the LLM-driven mapping function that contextualizes feature values within their statistical distributions. Table~\ref{tbl:dist} shows an example of the narrative profiling prompt, where the input includes raw feature values with exact numbers. Figure~\ref{fig:profile_comparison} illustrates the generated narrative profiles, where numerical features are described using relative trends and contextualized language.

\textbf{Historical Case Discovery.}
For a target user feature vector $\mathbf{x}_\theta$, we compute cosine similarity with each user feature vector $\mathbf{x}_j$ in the training set as $\mathcal{S}(\mathbf{x}_\theta, \mathbf{x}_j) = \frac{\mathbf{x}_\theta \cdot \mathbf{x}_j}{\|\mathbf{x}_\theta\| \|\mathbf{x}_j\|}$.

We first select the top-$\eta_1$ most similar users to compute mutual information (MI) for feature importance weighting. Then, we select the top-$\eta_2$ users ($\eta_2 < \eta_1$) as the reference set for causal structure learning using FCI.\footnote{Top-$\eta_2$ users are selected to balance FCI speed and performance.} From this set, we select representative cases $\mathcal{H}$ for few-shot reasoning, forming the foundation of \textbf{Pattern Analysis}.

\textbf{Causal Structure Learning.}
On the reference set (top-$\eta_2$ users), we apply the Fast Causal Inference (FCI) algorithm \cite{spirtes1995causal} to discover causal relationships between features $\mathbb{F}$ and the target variable $y \in \{0,1\}$. To quantify the importance of each causal feature $f_i \in \mathbb{F}$ connected to $y$, we leverage the previously computed mutual information scores. We select the top-$p$ causal features as $\mathcal{F}_c$, ensuring downstream reasoning focuses on key behavioral drivers.

These causal features provide focused guidance for the subsequent reasoning process and are integrated into the structured reasoning prompt as the basis for \textbf{Factor Analysis}.

\textbf{Structured Reasoning Framework.}
The structured reasoning prompt in AdaRec combines four components: the task description $T$, \textbf{Factor Analysis} (from causal features $\mathcal{F}_c$), \textbf{Pattern Analysis} (from representative cases $\mathcal{H}$), and the narrative profile $\Phi(\mathbf{x}_\theta)$. These elements guide the LLM to generate personalized and explainable recommendations: $R = \text{LLM}\Big(\xi\big(T, \mathcal{F}_c, \mathcal{H}, \Phi(\mathbf{x}_\theta)\big)\Big)$, where $\xi(\cdot)$ constructs a coherent reasoning input, and $R$ denotes the generated recommendation output. Table~\ref{tbl:brand_recommendation} shows an example of the structured reasoning prompt.

%% file: docs/4_experiment.tex
\section{Experiments}

\subsection{Tasks}
We evaluate AdaRec on two tasks: Customer Response Prediction (binary classification) and Brand Recommendation (top-3 selection from 17 brands). For Customer Response Prediction, the dataset contains 5,000 training samples, 600 validation samples, and 600 test samples, each with 119 features (115 numeric, 4 categorical). For Brand Recommendation, we use 4,692 training samples, 587 validation samples, and 587 test samples, each with 111 features (103 numeric, 8 categorical).

\subsection{Evaluation Metrics}
We report Precision, Recall, and F1 score for customer response prediction. For brand recommendation, we adopt Expected CTR \cite{li2011unbiased}, which measures the overlap between predicted and ground truth brand sets, weighted by click behaviour.

\subsection{Models}

\textbf{Baselines.}  
- \textbf{LightGBM} \cite{ke2017lightgbm}: Trained on 1.14M historical samples with 119 features for customer response prediction.\footnote{ML baselines use >1M samples; LLM-based models use fewer due to inference cost.}
- \textbf{Hierarchical RNN} \cite{du2015hierarchical}: For brand recommendation, trained on 2B+ user-brand historical interactions with 111 features.
- \textbf{MINT} \cite{mysore2023large}: Leverages LLMs to generate synthetic narrative queries from user-item interaction, enabling retrieval-based narrative-driven recommendations.
- \textbf{NBCRS} \cite{xie2024neighborhood}: Recommends by reusing crowd-written answers linked to similar conversational contexts, combining neighbourhood-based retrieval for an efficient conversational recommendation.
- \textbf{RecMind} \cite{wang2024recmind}: An autonomous LLM-powered agent that utilizes external knowledge and a self-inspiring planning algorithm to provide personalized recommendations.
\textbf{Expert Profiling}: Manually designed customer profiles by marketing experts based on 26 selected key features, providing static and quantitative summaries.

Figure~\ref{fig:profile_comparison} highlights the difference between expert profiling and AdaRec's narrative profiling.

\textbf{Backbone Models.}  
AdaRec is evaluated using Claude-3.5-Sonnet, Llama-3.1-70B, and Qwen-2.5-32B with greedy decoding (temperature = 0) and a maximum of 1000 output tokens.

\textbf{Parameters.}  
We set $k=5$ for the number of representative cases used in few-shot reasoning. We set $\eta_1=2000$, $\eta_2=1000$, use a significance level of $\alpha=0.1$, and select up to $p=15$ causal features in the causal analysis phase. To evaluate AdaRec's adaptability across different tasks, we apply supervised fine-tuning (SFT) for the Customer Response Prediction task (2 epochs, learning rate = $1\times10^{-4}$) and Kahneman-Tversky Optimization (KTO) for the Brand Recommendation task (2 epochs, learning rate = $5\times10^{-5}$). All experiments are conducted on 4 NVIDIA V100 GPUs.

\subsection{Results}

\textbf{RQ1: How effective is our approach compared to baselines?}  
Table~\ref{tbl:rq1} shows that AdaRec consistently outperforms all baselines across both tasks and profiling strategies. On the Customer Response Prediction task, AdaRec with Claude3.5 achieves an F1 score of 94.33\%, exceeding the strongest ML baseline (LightGBM at 86.67\%) by \textbf{8\%}. For the Brand Recommendation task, AdaRec with Qwen2.5 reaches a CTR of 10.3\%, outperforming all baselines.

In zero-shot settings, our narrative profiling shows significant advantages. For example, using Qwen2.5, the narrative profiling approach achieves an F1 score of 74.13\%, clearly surpassing expert profiling (55.58\%) with a \textbf{19\%} relative improvement.

Meanwhile, AdaRec performs well with both expert-designed and LLM-generated narrative profiles. In the simpler Customer Response Prediction task, Narrative + AdaRec achieves performance close to Expert + AdaRec (e.g., 91\% vs. 94\% F1 with Claude3.5). In contrast, for the more challenging Brand Recommendation task, Narrative + AdaRec outperforms Expert + AdaRec, with Qwen2.5 achieving a CTR of 10.3\% compared to 9.5\%. These results indicate that narrative profiling not only reduces manual feature engineering but also enhances performance in complex recommendation scenarios.

\begin{table}[h]
\begin{adjustbox}{width=\linewidth}
\begin{tabular}{lcc|ccc|c}
\hline
\multirow{2}{*}{Model} & \multirow{2}{*}{strategy} &\multirow{2}{*}{n-shot} & \multicolumn{3}{c|}{CustomerResponse} &BrandRec \\
 &  &  &P&R&F1    & CTR \\
\hline
\multicolumn{3}{c|}{ML Baseline} 
&86.67 & 86.67 & 86.67 &8.57 \\

\multicolumn{3}{c|}{MINT} 
&74.83 & 74.84 & 74.83 &8.4 \\

\multicolumn{3}{c|}{NBCRS} 
&84.01 & 84.01 & 84 &8.9 \\

\multicolumn{3}{c|}{RecMind} 
&79.25 & 78.62 & 78.54 &9 \\
\hline
\multirow{2}{*}{Llama3.1} 
& expert & 0 & 70.30 & 55.29 & 44.98 & 8.1 \\
& AdaRec(expert)  & 5 & 90.01 & 88.57 & 88.40 & 8.3 \\

& narrative 
& 0 & 75.40 & 68.18 & 65.62 & 8.1 \\

& AdaRec(narrative) & 5 & 90.38 & 90.03 & 89.98 & 8.5 \\

\multirow{2}{*}{Qwen2.5} 
& expert & 0 & 74.53 & 61.58 & 55.58 & 9.2 \\
& AdaRec(expert) & 5 & 92.72 & 92.66 & 92.66 & 9.5 \\

& narrative & 0 & 74.36 & 74.20 & 74.13 & 9.8 \\
& AdaRec(narrative)& 5 & 91.65 & 90.83 & 90.91 & \textbf{10.3} \\

\multirow{2}{*}{Claude3.5}  
& expert & 0 & 75.57 & 63.40 & 58.30 & 8.3 \\
& AdaRec(expert)& 5 & \textbf{94.42} & \textbf{94.35} & \textbf{94.33} & 8.3 \\

& narrative & 0 & 77.56 & 74.12 & 73.19 & 8.4 \\
& AdaRec(narrative) & 5 & 91.25 & 91.03 & 90.99 & 8.7 \\
\hline
\end{tabular}
\end{adjustbox}
\caption{Performance Comparison of Different Models and Strategies. All metrics are shown in percentages (\%). The best results are in bold.}
\label{tbl:rq1}
\vspace{-0.2in}
\end{table}

\textbf{RQ2: Can AdaRec generalize across tasks without retraining?}
To assess AdaRec’s ability to capture transferable customer representations, we conduct a cross-task evaluation: fine-tuning the model on one task (e.g., Customer Response Prediction), then directly applying it to another task (e.g., Brand Recommendation).

As shown in Table~\ref{tbl:rq2}, AdaRec maintains strong performance across tasks. In the zero-shot setting, the F1 score on Customer Response Prediction drops only slightly (from 75.7\% to 73.5\%), and Brand Recommendation CTR remains stable (9.0\% vs. 9.1\%). In the 5-shot setting, cross-task results closely match the in-task upper bound: 87.4\% vs. 87.6\% F1 and 9.7\% vs. 9.7\% CTR.

These results show that AdaRec learns generalizable representations of customer behavior that transfer well across tasks. The narrative profile supports robust reasoning across settings, demonstrating its potential to eliminate the need for repeated training and feature redesign in multi-objective customer modeling.

\begin{table}[h]
\centering
\begin{adjustbox}{width=\linewidth}
\begin{tabular}{lcc|ccc|c}
\hline
\multirow{2}{*}{Setting} & \multirow{2}{*}{Model} & \multirow{2}{*}{n-shot} & \multicolumn{3}{c|}{CustomerResponse (F1)} & BrandRec (CTR) \\
 & & & P & R & F1 & CTR \\
\hline
\multicolumn{7}{l}{\textit{Same-task evaluation (fine-tuned and tested on the same task)}} \\
\hline
In-task & Qwen2.5 & 0 & 75.76 & 75.69 & 75.65 & 9.0 \\
In-task & Qwen2.5 & 5 & 88.44 & 87.62 & 87.59 & 9.7 \\
\hline
\multicolumn{7}{l}{\textit{Cross-task evaluation (fine-tuned on one task, tested on another)}} \\
\hline
Cross-task & Qwen2.5 & 0 & 73.72 & 73.53 & 73.45 & 9.1 \\
Cross-task & Qwen2.5 & 5 & 88.37 & 87.44 & 87.40 & 9.7 \\
\hline
\end{tabular}
\end{adjustbox}
\caption{Cross-task generalization: AdaRec is fine-tuned on one task and directly evaluated on another. Results show minimal performance degradation.}
\label{tbl:rq2}
\vspace{-0.1in}
\end{table}

\textbf{RQ3: How robust is AdaRec to variations in narrative profiles?}
We evaluate AdaRec's robustness by using narrative profiles generated by different LLMs (Claude3.5, Llama3.1, Qwen2.5). As shown in Table~\ref{tbl:rq3}, AdaRec consistently delivers strong performance regardless of profiling source.

In the zero-shot setting, Customer Response Prediction F1 scores remain stable (71.9\% to 74.9\%), with Qwen2.5-generated profiles achieving the highest CTR (9.8\%) for Brand Recommendation.

In the 5-shot setting, AdaRec achieves over 90\% F1 across all profiling sources, with Claude3.5 profiles slightly outperforming others in Customer Response Prediction. CTR values remain competitive (8.9\% to 9.8\%), confirming AdaRec's ability to effectively leverage diverse narrative profiles.

These results demonstrate that AdaRec is robust to variations in narrative profile generation, ensuring reliable performance across different LLM providers and adaptable to various business environments.

\begin{table}[h]
\centering
\begin{adjustbox}{width=\linewidth}
\begin{tabular}{l|c|c|c}
\hline
\textbf{Profiling Source} & \textbf{CustomerResponse (F1)} & \textbf{BrandRec (CTR)} & \textbf{n-shot} \\
\hline
Claude3.5   & 74.9  / 94.4  & 8.5  / 9.8  & 0 / 5 \\
Llama3.1    & 71.9  / 93.9  & 8.1  / 8.9  & 0 / 5 \\
Qwen2.5     & 74.1  / 90.9  & 9.8  / 9.8  & 0 / 5 \\
\hline
\end{tabular}
\end{adjustbox}
\caption{AdaRec performance using narrative profiles from different LLMs. Metrics are shown as 0-shot / 5-shot.}
\label{tbl:rq3}
\vspace{-0.15in}
\end{table}

\vspace{-0.1in}
\begin{figure}[h]
\centering
\includegraphics[width=6.5cm]{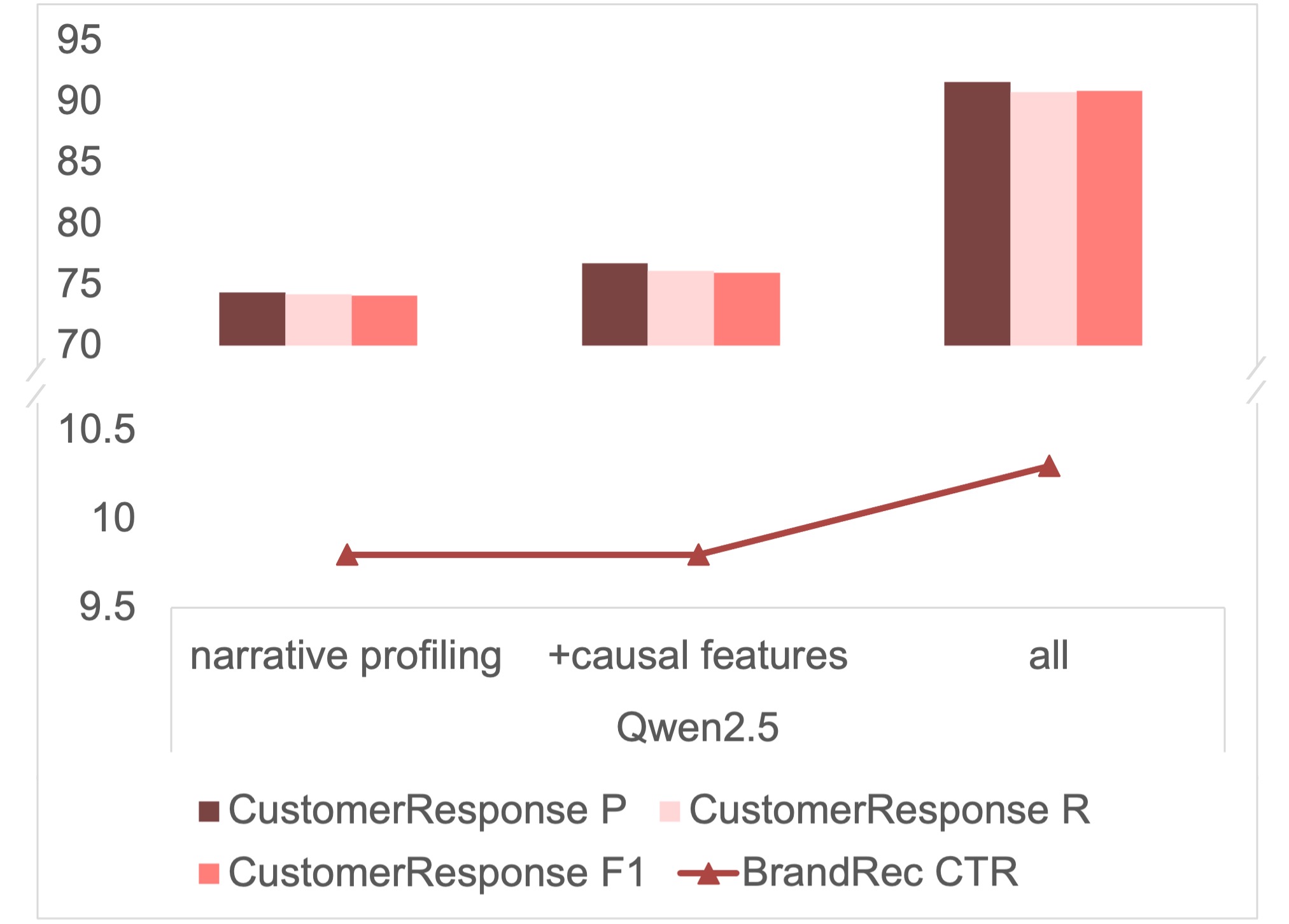}
\caption{The ablation study of AdaRec.}
\vspace{-0.15in}
\label{fig:ablation}
\end{figure}

\textbf{RQ4: How do different components contribute to AdaRec's performance?}
An ablation study (Figure~\ref{fig:ablation}) shows that narrative profiling provides a solid baseline (F1: 74.1\%, CTR: 9.8\%), causal feature weights offer moderate gains (F1: 76.0\%, CTR: 9.8\%), and historical behavior patterns deliver the largest improvement (F1: 90.9\%, CTR: 10.3\%). These results highlight the key role of historical patterns and the synergy of all components in achieving AdaRec’s high performance.

%% file: docs/5_conclusion.tex
\section{Conclusion}
This paper presents AdaRec, an adaptive recommendation framework that leverages LLMs to generate narrative profiles from user-item interactions for flexible and explainable personalization. Its dual-channel architecture, combining causal reasoning and similarity retrieval, delivers strong zero-shot and few-shot performance while reducing manual feature engineering and training costs. Experiments on real e-commerce datasets demonstrate notable improvements over ML models and LLM-based baselines. AdaRec supports real-time preference adaptation and explainable recommendations, making it practical for dynamic environments. Future work will focus on online A/B testing and extending to multi-modal tasks to enhance performance in real business applications.

%% file: custom.bib
@inproceedings{he2017neural,
  title={Neural collaborative filtering},
  author={He, Xiangnan and Liao, Lizi and Zhang, Hanwang and Nie, Liqiang and Hu, Xia and Chua, Tat-Seng},
  booktitle={Proceedings of the 26th international conference on world wide web},
  pages={173--182},
  year={2017}
}

@inproceedings{lin2024data,
  title={Data-efficient Fine-tuning for LLM-based Recommendation},
  author={Lin, Xinyu and Wang, Wenjie and Li, Yongqi and Yang, Shuo and Feng, Fuli and Wei, Yinwei and Chua, Tat-Seng},
  booktitle={Proceedings of the 47th international ACM SIGIR conference on research and development in information retrieval},
  pages={365--374},
  year={2024}
}

@article{zhang2023recommendation,
  title={Recommendation as instruction following: A large language model empowered recommendation approach},
  author={Zhang, Junjie and Xie, Ruobing and Hou, Yupeng and Zhao, Xin and Lin, Leyu and Wen, Ji-Rong},
  journal={ACM Transactions on Information Systems},
  year={2023},
  publisher={ACM New York, NY}
}

@inproceedings{bao2023tallrec,
  title={Tallrec: An effective and efficient tuning framework to align large language model with recommendation},
  author={Bao, Keqin and Zhang, Jizhi and Zhang, Yang and Wang, Wenjie and Feng, Fuli and He, Xiangnan},
  booktitle={Proceedings of the 17th ACM Conference on Recommender Systems},
  pages={1007--1014},
  year={2023}
}

@inproceedings{sheng2021one,
  title={One model to serve all: Star topology adaptive recommender for multi-domain ctr prediction},
  author={Sheng, Xiang-Rong and Zhao, Liqin and Zhou, Guorui and Ding, Xinyao and Dai, Binding and Luo, Qiang and Yang, Siran and Lv, Jingshan and Zhang, Chi and Deng, Hongbo and others},
  booktitle={Proceedings of the 30th ACM International Conference on Information \& Knowledge Management},
  pages={4104--4113},
  year={2021}
}

@inproceedings{pi2020search,
  title={Search-based user interest modeling with lifelong sequential behavior data for click-through rate prediction},
  author={Pi, Qi and Zhou, Guorui and Zhang, Yujing and Wang, Zhe and Ren, Lejian and Fan, Ying and Zhu, Xiaoqiang and Gai, Kun},
  booktitle={Proceedings of the 29th ACM International Conference on Information \& Knowledge Management},
  pages={2685--2692},
  year={2020}
}

@inproceedings{pi2019practice,
  title={Practice on long sequential user behavior modeling for click-through rate prediction},
  author={Pi, Qi and Bian, Weijie and Zhou, Guorui and Zhu, Xiaoqiang and Gai, Kun},
  booktitle={Proceedings of the 25th ACM SIGKDD international conference on knowledge discovery \& data mining},
  pages={2671--2679},
  year={2019}
}

@inproceedings{zhou2019deep,
  title={Deep interest evolution network for click-through rate prediction},
  author={Zhou, Guorui and Mou, Na and Fan, Ying and Pi, Qi and Bian, Weijie and Zhou, Chang and Zhu, Xiaoqiang and Gai, Kun},
  booktitle={Proceedings of the AAAI conference on artificial intelligence},
  volume={33},
  number={01},
  pages={5941--5948},
  year={2019}
}

@inproceedings{zhou2018deep,
  title={Deep interest network for click-through rate prediction},
  author={Zhou, Guorui and Zhu, Xiaoqiang and Song, Chenru and Fan, Ying and Zhu, Han and Ma, Xiao and Yan, Yanghui and Jin, Junqi and Li, Han and Gai, Kun},
  booktitle={Proceedings of the 24th ACM SIGKDD international conference on knowledge discovery \& data mining},
  pages={1059--1068},
  year={2018}
}

@inproceedings{mysore2023large,
  title={Large language model augmented narrative driven recommendations},
  author={Mysore, Sheshera and McCallum, Andrew and Zamani, Hamed},
  booktitle={Proceedings of the 17th ACM Conference on Recommender Systems},
  pages={777--783},
  year={2023}
}

@article{wang2023recmind,
  title={Recmind: Large language model powered agent for recommendation},
  author={Wang, Yancheng and Jiang, Ziyan and Chen, Zheng and Yang, Fan and Zhou, Yingxue and Cho, Eunah and Fan, Xing and Huang, Xiaojiang and Lu, Yanbin and Yang, Yingzhen},
  journal={arXiv preprint arXiv:2308.14296},
  year={2023}
}

@article{weng2004feature,
  title={Feature-based recommendations for one-to-one marketing},
  author={Weng, Sung-Shun and Liu, Mei-Ju},
  journal={Expert Systems with Applications},
  volume={26},
  number={4},
  pages={493--508},
  year={2004},
  publisher={Elsevier}
}

@inproceedings{wang2024recmind,
  title={RecMind: Large Language Model Powered Agent For Recommendation},
  author={Wang, Yancheng and Jiang, Ziyan and Chen, Zheng and Yang, Fan and Zhou, Yingxue and Cho, Eunah and Fan, Xing and Lu, Yanbin and Huang, Xiaojiang and Yang, Yingzhen},
  booktitle={Findings of the Association for Computational Linguistics: NAACL 2024},
  pages={4351--4364},
  year={2024}
}

@inproceedings{xie2024neighborhood,
  title={Neighborhood-Based Collaborative Filtering for Conversational Recommendation},
  author={Xie, Zhouhang and Wu, Junda and Jeon, Hyunsik and He, Zhankui and Steck, Harald and Jha, Rahul and Liang, Dawen and Kallus, Nathan and McAuley, Julian},
  booktitle={Proceedings of the 18th ACM Conference on Recommender Systems},
  pages={1045--1050},
  year={2024}
}

@inproceedings{li2011unbiased,
  title={Unbiased offline evaluation of contextual-bandit-based news article recommendation algorithms},
  author={Li, Lihong and Chu, Wei and Langford, John and Wang, Xuanhui},
  booktitle={Proceedings of the fourth ACM international conference on Web search and data mining},
  pages={297--306},
  year={2011}
}

@inproceedings{du2015hierarchical,
  title={Hierarchical recurrent neural network for skeleton based action recognition},
  author={Du, Yong and Wang, Wei and Wang, Liang},
  booktitle={Proceedings of the IEEE conference on computer vision and pattern recognition},
  pages={1110--1118},
  year={2015}
}

@article{ke2017lightgbm,
  title={Lightgbm: A highly efficient gradient boosting decision tree},
  author={Ke, Guolin and Meng, Qi and Finley, Thomas and Wang, Taifeng and Chen, Wei and Ma, Weidong and Ye, Qiwei and Liu, Tie-Yan},
  journal={Advances in neural information processing systems},
  volume={30},
  year={2017}
}

@inproceedings{spirtes1995causal,
  title={Causal inference in the presence of latent variables and selection bias},
  author={Spirtes, Peter and Meek, Christopher and Richardson, Thomas},
  booktitle={Proceedings of the Eleventh conference on Uncertainty in artificial intelligence},
  pages={499--506},
  year={1995}
}

@inproceedings{hansen2020contextual,
  title={Contextual and sequential user embeddings for large-scale music recommendation},
  author={Hansen, Casper and Hansen, Christian and Maystre, Lucas and Mehrotra, Rishabh and Brost, Brian and Tomasi, Federico and Lalmas, Mounia},
  booktitle={Proceedings of the 14th ACM Conference on Recommender Systems},
  pages={53--62},
  year={2020}
}

@inproceedings{wang2015collaborative,
  title={Collaborative deep learning for recommender systems},
  author={Wang, Hao and Wang, Naiyan and Yeung, Dit-Yan},
  booktitle={Proceedings of the 21th ACM SIGKDD international conference on knowledge discovery and data mining},
  pages={1235--1244},
  year={2015}
}

@article{wu2024survey,
  title={A survey on large language models for recommendation},
  author={Wu, Likang and Zheng, Zhi and Qiu, Zhaopeng and Wang, Hao and Gu, Hongchao and Shen, Tingjia and Qin, Chuan and Zhu, Chen and Zhu, Hengshu and Liu, Qi and others},
  journal={World Wide Web},
  volume={27},
  number={5},
  pages={60},
  year={2024},
  publisher={Springer}
}

@inproceedings{zhang2024notellm,
  title={NoteLLM: A Retrievable Large Language Model for Note Recommendation},
  author={Zhang, Chao and Wu, Shiwei and Zhang, Haoxin and Xu, Tong and Gao, Yan and Hu, Yao and Chen, Enhong},
  booktitle={Companion Proceedings of the ACM on Web Conference 2024},
  pages={170--179},
  year={2024}
}
